\PassOptionsToPackage{table}{xcolor}
\documentclass[11pt]{article}

\usepackage[final]{acl}

\usepackage{enumitem}
\usepackage{xcolor}
\definecolor{grey}{rgb}{0.5,0.5,0.5}
\usepackage{times}
\usepackage{latexsym}

\usepackage[T1]{fontenc}

\usepackage[utf8]{inputenc}

\usepackage{microtype}

\usepackage{inconsolata}

\usepackage{graphicx}

\usepackage{xspace}
\usepackage{verbatim}
\usepackage{todonotes}
\usepackage{amsmath}
\usepackage{amsfonts}
\usepackage{enumitem}
\usepackage{cleveref}
\usepackage{stfloats}  
\usepackage{booktabs}
\usepackage{subcaption}

\usepackage{setspace}

\usepackage{tcolorbox}
\tcbset{colback=gray!10, colframe=black!50, boxrule=0.5pt, arc=2pt, left=2mm, right=2mm, top=1mm, bottom=1mm}

\usepackage{graphicx}
\usepackage{xcolor}

\usepackage{soul}

\title{Does The Way You Plan Matter?\\ An Empirical Study of Planning Representations for LLM Web Agents}

\author{
 \textbf{Alejandra Zambrano\textsuperscript{1,2}},
 \textbf{Sara Vera Marjanović\textsuperscript{2,3}},
 \textbf{Imene Kerboua\textsuperscript{4}},
 \textbf{Xing Han Lù \textsuperscript{2,5}},
\\
 \textbf{Leila Kosseim\textsuperscript{1}},
\\
 \textsuperscript{1}Concordia University,
 \textsuperscript{2}Mila - Quebec AI Institute 
\\
 \textsuperscript{3}University of Copenhagen,
 \textsuperscript{4}Universite Claude Bernard Lyon,
 \textsuperscript{5}McGill University,
\\
 \small{
   \textbf{Correspondence:} \href{mailto:al_zambr@live.concordia.ca}{al\_zambr@live.concordia.ca}
 }
}

\newcommand{\method}{\textsc{PlanAhead}\xspace}

\begin{document}
\maketitle
\begin{abstract}



Despite recent advances, LLM-based web agents still struggle with limited exploration, omission of critical steps, and sensitivity to task constraints. Prior work suggests that many of these failures stem from weaknesses in planning, yet the impact of alternative natural language plan representations remains unexplored. To address this, we introduce \method, a static planner-executor framework that evaluates the impact of plan representation on agent performance. We first automatically categorize WebArena tasks into 3 difficulty levels, enabling consistent difficulty grading without human annotation. Then we systematically evaluate 4 different plan representations on the tasks categorized as hard: sequential subgoals, narrative, pseudocode, and checklist; across different families of multimodal LLM-powered agents(OpenAI, Alibaba, and Google). To account for stochastic variability, we introduce a novel decomposition of agent Success Rate (SR) into 2 interpretable evaluation metrics: Achievement Rate (AR) and Solved-Task Consistency (STC). Our results show that both, the plan formulation and the underlying LLM generating the plan, significantly influence web-agent robustness and task success. Code and dataset are available \href{https://github.com/alzambranolu13/plan-ahead}{here}.

\end{abstract}

\section{Introduction}
\label{sec:introduction}

Recent advances in large language models (LLMs) have substantially enhanced the capabilities of web agents. Yet, web tasks, where agents must navigate the Internet to fulfill user-defined goals~\citep{deng2023mind2web,zhou2023webarena}, remain particularly challenging, as LLM-based web agents still struggle with adaptability, real-time interactions, and robustness in dynamic contexts~\cite{chae2024web,xue2025illusion}. 

Planning has emerged as a key strategy for improving web-agent performance, allowing complex tasks to be decomposed into simpler steps~\citep{prasad2023adapt,sodhi2023step}. While most existing approaches express plans as sequential lists of subgoals, recent work has explored alternative representations beyond natural language~\citep{wei2025plangenllms}. However, alternative natural language (NL) representations of plans remain largely unexplored.

In this paper, we challenge both the definition of task difficulty and the representation of plans. Rather than proposing a new planning algorithm, we conduct an empirical study of how different plan representations affect web-agent performance and robustness. We introduce \method, a framework that redefines difficulty grading and systematically evaluates diverse plan formats across multiple multimodal LLMs from different families (OpenAI, Alibaba, and Google). The contributions are threefold:
\begin{enumerate}[noitemsep,nolistsep]   
    \item \textbf{Automated task difficulty grading.} We introduce a reproducible pipeline for partitioning benchmark tasks into 3 difficulty levels: \textit{Easy}, \textit{Medium}, and \textit{Hard} without the need for human annotation. This categorization enables more systematic evaluation of planning strategies under varying levels of task complexity.  
    \item \textbf{Evaluation of novel planning representations.} We propose 3 alternative NL-based planning formats: narrative, pseudocode, and requirement checklists, and compare them against sequential subgoal decomposition. Our findings reveal that non-standard formats can, in specific conditions, be beneficial for web agents, particularly on \textit{Hard} tasks. 
    \item \textbf{Novel interpretable decomposition of Success Rate (SR)}. We decompose SR into \textit{Achievement Rate} (AR) and \textit{Solved-Task Consistency} (STC). AR, related to empirical pass@k, measures the proportion of tasks solved at least once across repeated runs, acknowledging that agents may not succeed on every attempt. STC measures the repeatability of success among those achieved tasks. 
    
\end{enumerate}
\begin{figure*}[t]
    \centering
    \includegraphics[width=0.75\linewidth]{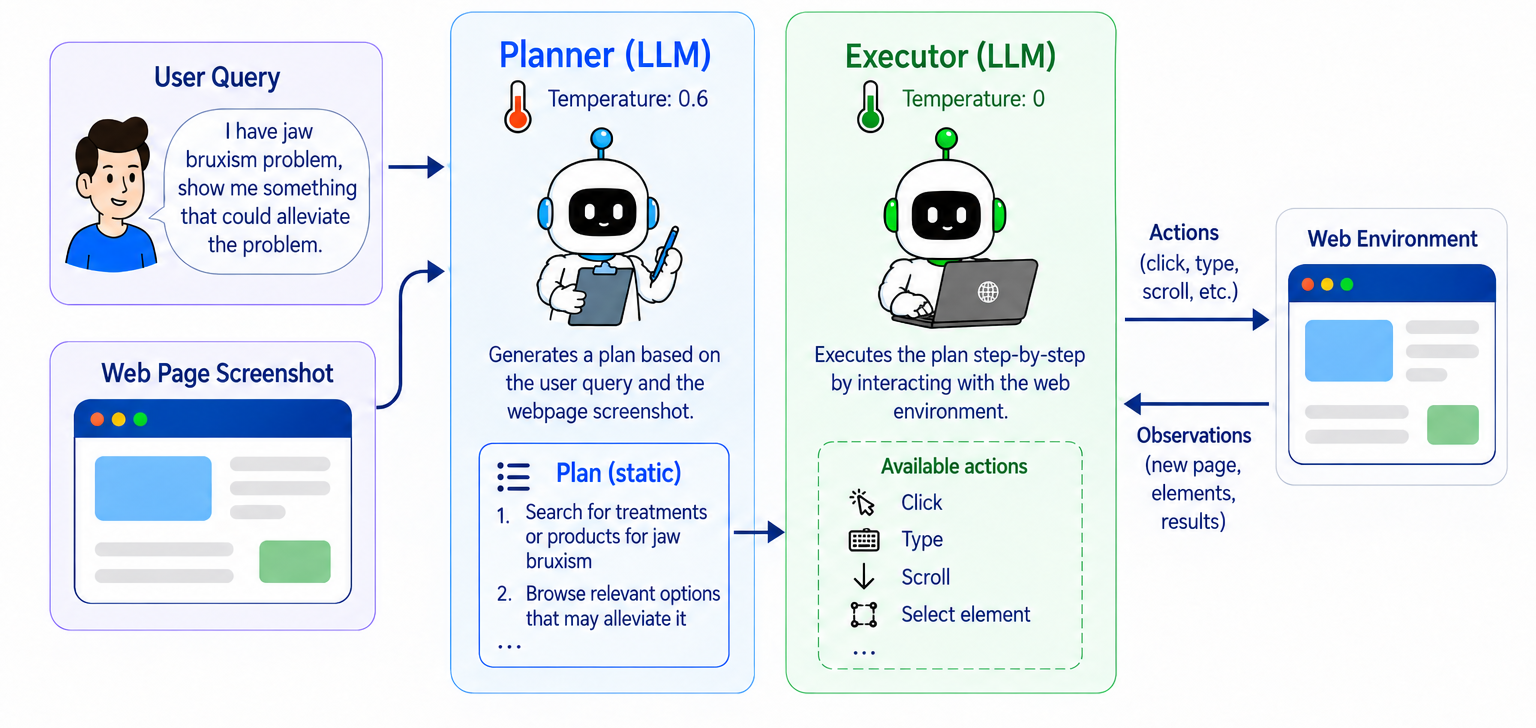}
    \caption{Overview of our framework \method}
    \label{fig:main_fig}
\end{figure*}

\section{Related Works}
\label{sec:related_works}
\paragraph{Planning in Web Agents.}Planning through task decomposition, where complex tasks are broken into simpler subgoals~\citep{huang2024understanding}, has emerged as a key capability for tackling complex, multi-step tasks in dynamic environments. Planning can be implemented within a single agent~\citep{sodhi2023step}, but more often through a planner-executor separation, where one LLM generates plans and another executes them~\citep{erdogan2025plan,abuelsaad2024agent,agashe2025agent,gur2023real,zhang2025planning,marreed2025towards,zhang2025webpilot}. This separation enables the planner to provide high-level structure, while the executor focuses on step-by-step execution.
Planners normally follow two types of approaches: static or dynamic. Static plans are generated only once and are not reevaluated during execution, while dynamic plans can be reevaluated either when the executor finds an issue~\citep{prasad2023adapt} (reactive planning) or at every step of the task execution~\citep{agashe2025agent} (proactive planning). 
Despite extensive research, studies show that LLMs still struggle to generate accurate and reliable plans~\citep{valmeekam2023planning}. In most existing approaches, plans are expressed as sequential lists of subgoals. However, exploring alternative plan representations, both different natural language forms and non–natural language formats, remains largely unexplored~\citep{wei2025plangenllms}. Understanding this is particularly important given that planning capabilities have a significant impact on web-agent success~\cite{shlomov2024grounding}.

\paragraph{Multi-Run Evaluation Metrics.}

Several metrics have been proposed for evaluating web agents, with \emph{Success Rate (SR)} remaining the most widely used~\citep{yehudai2025survey}. SR measures the proportion of tasks successfully completed out of the total number of evaluated tasks. However, most existing evaluations report performance from a single run per task, providing limited information about whether a task is consistently solvable or only occasionally achieved across different runs. A related idea appears in code-generation evaluation for LLMs through the \textit{pass@k} metric, introduced by~\cite{chen2021evaluating}. In this setting, multiple code samples are generated for each programming problem. A problem is considered solved if at least one of the generated samples passes the corresponding unit tests. The underlying intuition of pass@k is highly relevant for stochastic agents: a task may be achievable even if it is not solved in every attempt. However, this perspective remains underexplored in web-agent evaluation.

\paragraph{Web-Agent Benchmarks.} Several web-agent benchmarks have been proposed since~\cite{shi2017world}, but many introduce limitations for controlled evaluation. In particular, offline benchmarks such as Mind2Web~\citep{deng2023mind2web} and WebLINX~\citep{lu2024weblinx} are built from real-world interactions, but evaluate next-action prediction over recorded trajectories rather than overall goal completion. Live-web benchmarks such as Online-Mind2Web~\citep{xue2025illusion} provide this interactivity, but introduce reproducibility concerns because of the website's content and external dependencies may change over time. 
We therefore use WebArena~\citep{zhou2023webarena}, which provides a dynamic yet controlled environment for web-agent evaluation. Its sandbox setup preserves interactive task execution while supporting reproducible evaluation across a diverse set of challenging tasks.





\section{Methodology}
\label{sec:methods}
\begin{figure*}[t]
    \centering
    \includegraphics[width=0.85\linewidth]{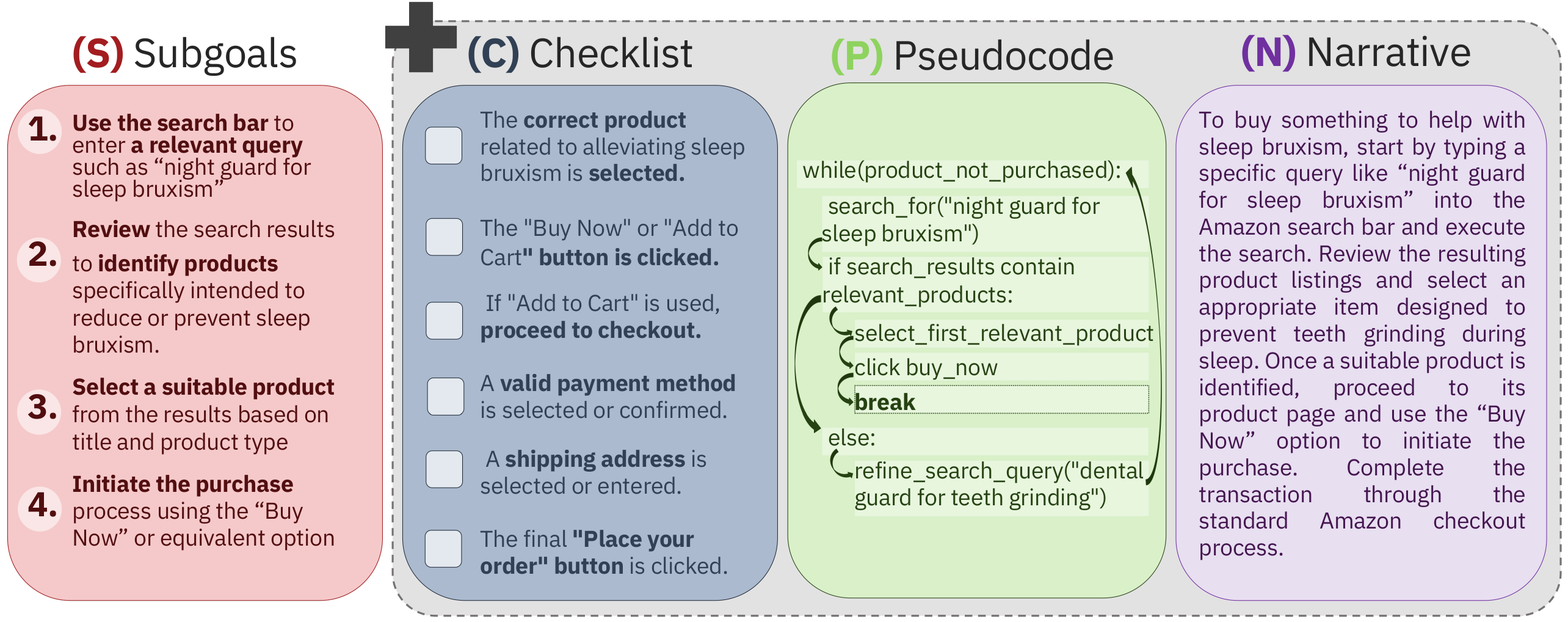}
    \caption{The four representations used in our evaluation of \method. On the left, we show the standard plan of sequential subgoals. In the box on the right, we show three new planning representations. All strategies were generated by GPT-4o given the task \textit{"Buy something to alleviate sleep bruxism"}. Details on prompt templates used to generate each representation are provided in Appendix~\ref{app:prompts}.}
    \label{fig:examples}
\end{figure*}




\subsection{Task Difficulty Categorization}
Our primary interest is in tasks that are not achievable by a non-planning agent, as these provide the clearest setting for measuring the added value of planning. Re-evaluating tasks that the baseline agent can already solve offers limited insight into whether planning itself enables new capabilities. Instead, we focus on identifying tasks that the baseline consistently fails and ask whether introducing an explicit planning module can make them achievable. To identify these challenging tasks in a reproducible manner, we start from the \textit{WebArena} test split provided through \textit{BrowserGym}~\citep{dechezelles2025browsergymecosystem}, which consists of 381 tasks. We partition these tasks into three empirical difficulty levels: \textit{Easy}, \textit{Medium}, and \textit{Hard}. Difficulty is estimated using BrowserGym's \emph{GenericAgent} as our baseline agent, a lightweight LLM-based web agent that receives structured browser observations (e.g., AXTree, HTML, and screenshots) and predicts the next action toward the task goal.

To reduce the dependence of the resulting labels on a single model or execution trajectory, we evaluate \emph{GenericAgent} with five different LLM backbones (GPT 4.1, GPT 4.1-mini, GPT 4o-mini, Gemini 2.5 Flash and Qwen 2.5 VL 72B) and run 5 independent trials per model. This results in 25 attempts per task. We set a temperature of 0.4 at each run to induce moderate variation across trajectories while maintaining stable agent behavior.

We assign a task to the \textit{Easy} category only if \textbf{all five models succeed in all five trials} (25/25 successful runs), and to the \textit{Hard} category only if \textbf{all five models fail in all five trials} (0/25 successful runs). All remaining tasks are labeled \textit{Medium}. This deliberately strict criterion makes the extreme categories conservative: \textit{Easy} tasks are those that are consistently solvable across models and runs, whereas \textit{Hard} tasks are those for which the baseline agent fails to find a successful trajectory across a diverse set of 25 attempts.

The resulting partition contains 14 \textit{Easy}, 209 \textit{Medium}, and 158 \textit{Hard} tasks. Our subsequent analysis focuses on the \textit{Hard} subset. Additional details and analyses of the task grading are provided in Appendix~\ref{app:hard_set}.

\subsection{Planner-Executor Modules}
\method follows a static planner-executor framework (see Section~\ref{sec:related_works}). First, an LLM planner generates a plan from the WebArena task goal and a screenshot of the initial browser state. This step is performed once at the beginning of the task, producing a static plan.
As shown in Figure \ref{fig:main_fig}, the plan is then provided to the executor LLM together with the current browser state, and the executor predicts low-level actions at each step. The planner is sampled with a temperature of 0.6 to encourage plan diversity, while the executor operates deterministically (temperature of 0) to ensure stable action selection. This setup ensures that performance differences arise from the plan representation rather than action-level stochasticity. The executor is allowed up to 30 actions per task, and success is evaluated using WebArena’s automatic reward.

As a dynamic planning baseline, we use the \emph{GenericAgent} \texttt{UsePlan} capability, which combines plan generation and action prediction by producing a new sequential plan at each step by a single agent.
\subsection{Planning Representations}

Using the 158 \textit{Hard} tasks, we evaluate 4 plan representations for communicating high-level plans to the executor: a standard sequential list of subgoals and 3 novel representations: Requirement Checklist, Pseudocode, and Narrative. These representations differ in abstraction level, information ordering, and the degree of flexibility given to the executor. Figure~\ref{fig:examples} illustrates each representation.







\subsection{Evaluation Metrics}
As discussed in Section~\ref{sec:related_works}, the stochasticity of LLM-based agents provides an opportunity to explore multiple trajectories toward the same goal. However, standard SR is typically defined over a single run per task and therefore does not capture this potential. When evaluating multiple runs, we instead want to distinguish between whether a task can be achieved and how consistently it can be achieved. We therefore decompose SR into two complementary metrics:


\paragraph{Achievement Rate (AR)} measures task-level achievability in a set of $T$ tasks over $R$ independent runs. A task is considered achieved if at least one run succeeds:
\begin{equation}
{\small
\mathrm{AR} = \frac{1}{T} \sum_{t=1}^{T} \left[\bigvee_{i=1}^{R} \mathrm{Reward}(t, i)\right] 
}
\end{equation}
Where $\mathrm{Reward}(t,i)$ is a binary indicator of whether task $t$ succeeds in run $i$.
\paragraph{Solved-Task Consistency (STC)} For tasks that were achieved at least once (i.e. $A = \{\, t \mid \exists i \in \{1,\dots,R\} : \mathrm{Reward}(t,i)=1 \,\}$), STC measures performance consistency across runs:
\begin{equation}
{\small
\mathrm{STC} = \frac{1}{|A| \times R} \sum_{t \in A}\sum_{i=1}^{R} \mathrm{Reward}(t,i) 
}
\label{eq:stc}
\end{equation}
Together, these metrics provide an exact decomposition of Success Rate $SR=AR \times STC$.

\paragraph{As an illustrative example,} consider three web tasks, each evaluated over five runs. Suppose that only one task succeeds, and that it succeeds in two out of its five runs, while the other two tasks fail in all runs. Using the standard SR, the 15 task attempts are aggregated together, giving $SR=\frac{2}{3 \times 5} = 13.3\%$. In contrast, AR captures achievability at the task level giving an $AR=\frac{1}{3} = 33.3\%$ because one out of the three tasks is achieved. STC then measures reliability among the achieved tasks resulting in $STC=\frac{2}{1\times5} = 40\%$. 



\section{Results and Analysis}
\label{sec:results}
\begin{table*}[ht!]
\centering
\resizebox{\linewidth}{!}{
\begin{tabular}{c cc|rr|rr|rr|rr|rr}
\hline
\multicolumn{3}{c|}{\textbf{Configuration}} 
& \multicolumn{2}{c|}{\textbf{Dynamic}} 
& \multicolumn{8}{c}{\textbf{Static (Ours)}} \\
\textbf{\#} & \textbf{Planner} & \textbf{Executor}
& \multicolumn{2}{c|}{\textbf{Sequential}}
& \multicolumn{2}{c|}{\textbf{Sequential}}
& \multicolumn{2}{c|}{\textbf{Checklist}}
& \multicolumn{2}{c|}{\textbf{Pseudocode}}
& \multicolumn{2}{c}{\textbf{Narrative}} \\
\hline
&&& \textbf{AR} & \textbf{STC}
& \textbf{AR} & \textbf{STC}
& \textbf{AR} & \textbf{STC}
& \textbf{AR} & \textbf{STC}
& \textbf{AR} & \textbf{STC} \\
\hline

1 & GPT 4.1-mini & GPT 4.1-mini
& 5.1 & 42
& 5.1 & 42
& 4.4 & 35
& 4.4 & 34
& \textbf{5.1} & \textbf{75} \\

2 & Qwen 2.5-VL 72B & GPT 4.1-mini
& - & -
& 5.4 & 31
& 4.2 & 40
& 5.4 & 31
& \textbf{6.6} & \textbf{60} \\

3 & Gemini 2.5 Flash& GPT 4.1-mini
& - & -
& \textbf{7.0} & 45
& 3.2 & 36
& 4.4 & 51
& 5.7 & \textbf{53} \\

4 & Qwen 2.5-VL 72B & Qwen 2.5-VL 72B
& 5.1 & 55
& 2.5 & 85
& \textbf{5.1} & \textbf{75}
& 3.2 & 80
& 3.8 & 63 \\

5 & GPT 4.1-mini & Qwen 2.5-VL 72B
& - & -
& 5.1 & 35
& \textbf{7.5} & \textbf{30}
& 5.0 & 45
& 5.1 & 45 \\

6 & Gemini 2.5 Flash & Qwen 2.5-VL 72B
& - & -
& 3.2 & 32
& \textbf{5.7} & \textbf{40}
& 3.8 & 40
& 4.4 & 40 \\

7 & Gemini 2.5 Flash & Gemini 2.5 Flash
& 9.5 & 39
& 5.7 & 53
& 7.6 & 58
& 8.2 & 43
& \textbf{7.6} & \textbf{73} \\

8 & \cellcolor{grey!25}{GPT 4.1-mini} & \cellcolor{grey!25}{Gemini 2.5 Flash}
& \cellcolor{grey!25}{-} & \cellcolor{grey!25}{-}
& \cellcolor{grey!25}{\textbf{10.7}} & \cellcolor{grey!25}{\textbf{47}}
& \cellcolor{grey!25}{8.9} & \cellcolor{grey!25}{44}
& \cellcolor{grey!25}{7.0} & \cellcolor{grey!25}{53}
& \cellcolor{grey!25}{9.5} & \cellcolor{grey!25}{43} \\

9 & Qwen 2.5-VL 72B & Gemini 2.5 Flash
& - & -
& \textbf{8.9} & \textbf{50}
& 7.0 & 58
& 4.4 & 77
& \textbf{8.9} & \textbf{50} \\

\bottomrule
\end{tabular}
}
\caption[Comparison of Achievement Rate (AR) and Solved-Task Consistency (STC)]
{Comparison of percentage AR and STC across planning strategies on the \textit{Hard} set. In bold are the best results (of AR and STC) per configuration and highlighted in grey is the best configuration overall}
\label{tab:main_table_AR}
\end{table*}
We evaluate planning strategies on the 158 \textit{Hard} tasks. For each task and strategy, we run $R=5$ independent trials to observe execution variability and distinguish tasks that are occasionally solvable from those that are consistently solvable.
In our experiments we use a subset of the LLM backbones used for the task categorization: GPT 4.1-mini, Qwen-2.5-VL-72B, and Gemini 2.5 Flash, evaluated as both planner and executor, as well as all planner-executor combinations. Table \ref{tab:main_table_AR} shows the performance measured using AR and STC percentages. We also include a dynamic planning baseline (see Section~\ref{sec:methods}) to compare static and dynamic planning under identical conditions. Key findings are summarized below:

\paragraph{Planning makes unachievable tasks achievable.}
Because the Hard subset consists exclusively of tasks that were not solved by any non-planning agent across all runs, any AR greater than zero reflects task completion attributable to the introduction of planning. Across all configurations and representations, $AR > 0$ (see Table~\ref{tab:main_table_AR}). This demonstrates that changing the planning strategy (regardless of the representation or configuration) allows agents to solve tasks that were consistently unsolved under the baseline setting. 

\begin{figure}[h]
    \centering
    \includegraphics[width=\linewidth]{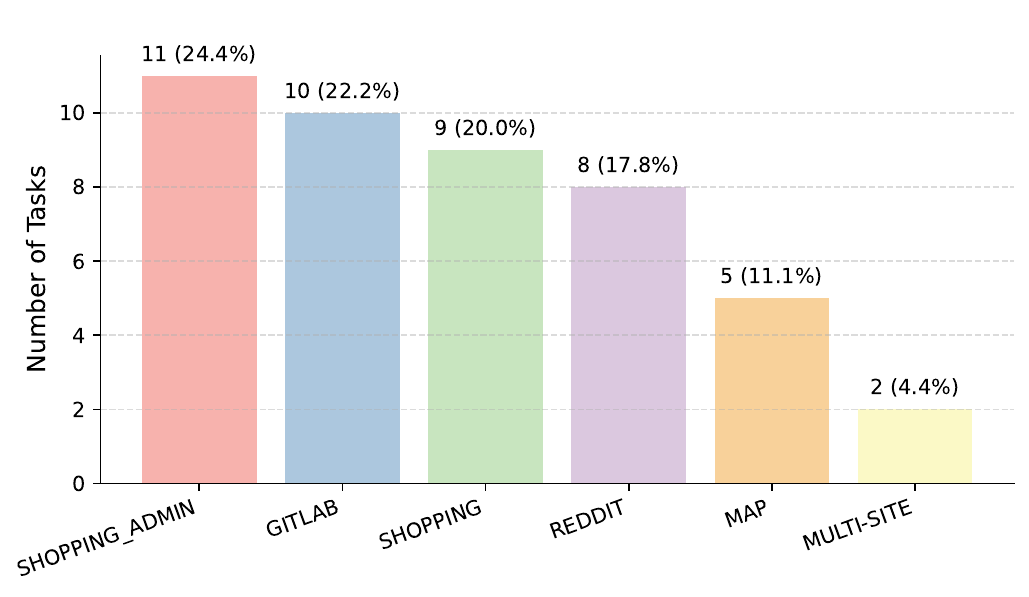}
    \caption{Task distribution of achieved tasks aggregating all types of planning.}
    \label{fig:achieved_task_dist}
\end{figure}

Figure~\ref{fig:achieved_task_dist} further illustrates these gains by showing the distribution of achieved tasks across \textit{WebArena} categories. Out of the 158 tasks classified as \textit{Hard}, 45 tasks were successfully completed (approximately 28\%) at least once by one or more of the planning configurations. These achieved tasks are distributed across all categories, suggesting that the benefits of planning are not limited to a single domain, but appear across a range of task types. 




\textbf{LLMs Benefit from Distinct Plan Representations.}
For example, when GPT 4.1-mini is used as executor, Narrative plans generally exhibit strong performance, achieving the highest AR in 2 configurations and particularly high STC across configurations (see rows 1,2 and 3 of Table~\ref{tab:main_table_AR}). A plausible explanation is that GPT-style models, trained extensively on natural language text, may better leverage coherent descriptive instructions when interpreting plans. In contrast, Qwen-2.5-VL performs best with checklist-based representations (see rows 4,5,6), which provide explicit and structured guidance. Gemini performs competitively across multiple representations and notably benefits from non-standard formats (see rows 7,8,9). 
Interestingly, Pseudocode, a representation not designed for human readability, yields one of the highest ARs observed when Gemini is used as both planner and executor (see row 7). This suggests that structured control flow may provide useful information for LLM reasoning, even when expressed in simplified pseudo-programmatic form.

Taken together, these results demonstrate that the effectiveness of planning is influenced by their representational form and that the traditional format of a sequential numbered list might not always be optimal.


\textbf{Planner-Executor Separation Reveals Complementary Model Strengths.}
Separating planning and execution exposes clear differences in model suitability for each role. Across configurations, mixed planner-executor pairs generally outperform homogeneous setups in AR. Notably, GPT 4.1-mini as planner combined with Gemini as executor achieves the strongest overall performance in our evaluation (see row 8 of Table~\ref{tab:main_table_AR}). While we typically see a drop in STC given an increase in AR (given~\ref{eq:stc}), this occurs \textit{without} a substantial increase in AR (compare row 4 and 6). This pattern indicates that some models, such as GPT 4.1-mini, are more effective at plan generation, whereas others, like Gemini, may be better suited for low-level action execution. These findings support modular agent architectures that exploit complementary model strengths and enable flexible model combinations to balance performance and cost.

\textbf{Static Planning is more Consistent.} Even without the possibility of regenerating or updating plans during execution, static plans achieve higher consistency (STC) than single-agent dynamic plans. Qualitative trace analysis shows that dynamic agents often enter action loops, lose track of task progress, or prematurely mark subtasks as completed (see Appendix~\ref{app:dynamic_plans}). Static planner-executor is a simple setup that allows the executor to focus exclusively on action selection.


\section{Conclusion}
\label{sec:conclusion}

In this work, we revisited the role of planning in web agents, challenging the assumption that plans must take the form of sequential subgoals. We introduced \method, an automated and reproducible approach for identifying challenging web tasks, and used it to conduct a controlled empirical study of planning representations with a new decomposition of SR into two interpretable metrics. Our experiments show that planning effectiveness depends not only on the models chosen but also on the representational form of the plan. Overall, our findings position planning as a crucial but flexible design dimension.


\section*{Limitations} Despite our comprehensive list of planning strategies, our study has several limitations. We restricted our evaluation to a small set of LLM backends (GPT-4.1-mini, Qwen-2.5-VL and Gemini 2.5 Flash), a single benchmark (WebArena), and a limited number of runs ($R=5$). Expanding to a broader range of models, tasks, and evaluation scales would provide a more comprehensive picture of planning effectiveness.

\bibliography{custom}

@inproceedings{zhou2023webarena,
    title     = {{WebArena}: A Realistic Web Environment for Building Autonomous Agents},
    author    = {Shuyan Zhou and Frank F. Xu and Hao Zhu and Xuhui Zhou and Robert Lo and Abishek Sridhar and Xianyi Cheng and Tianyue Ou and Yonatan Bisk and Daniel Fried and Uri Alon and Graham Neubig},
    booktitle = {Proceedings of the Twelfth International Conference on Learning Representations (ICLR)},
    year      = {2023},
    url       = {https://openreview.net/forum?id=oKn9c6ytLx}
}

@article{abuelsaad2024agent,
    title     = {Agent-e: From autonomous web navigation to foundational design principles in agentic systems},
    author    = {Abuelsaad, Tamer and Akkil, Deepak and Dey, Prasenjit and Jagmohan, Ashish and Vempaty, Aditya and Kokku, Ravi},
    journal   = {arXiv preprint arXiv:2407.13032},
    year      = {2024},
    url       = {https://arxiv.org/abs/2407.13032}
}

@article{prasad2023adapt,
    title     = {{ADAPT}: As-needed decomposition and planning with language models},
    author    = {Prasad, Archiki and Koller, Alexander and Hartmann, Mareike and Clark, Peter and Sabharwal, Ashish and Bansal, Mohit and Khot, Tushar},
    journal   = {arXiv preprint arXiv:2311.05772},
    year      = {2023},
    url       = {https://arxiv.org/abs/2311.05772}
}

@inproceedings{zhang2025webpilot,
    title     = {{WebPilot}: a versatile and autonomous multi-agent system for web task execution with strategic exploration},
    author    = {Zhang, Yao and Ma, Zijian and Ma, Yunpu and Han, Zhen and Wu, Yu and Tresp, Volker},
    booktitle = {Proceedings of the Thirty-Ninth AAAI Conference on Artificial Intelligence (AAAI-25)},
    year      = {2025},
    doi       = {10.1609/aaai.v39i22.34505},
    url       = {https://doi.org/10.1609/aaai.v39i22.34505}
}

@article{sodhi2023step,
    title     = {{SteP}: Stacked {LLM} policies for web actions},
    author    = {Sodhi, Paloma and Branavan, SRK and Artzi, Yoav and McDonald, Ryan},
    journal   = {arXiv preprint arXiv:2310.03720},
    year      = {2023},
    url       = {https://arxiv.org/abs/2310.03720}
}

@article{shlomov2024grounding,
    title     = {From grounding to planning: Benchmarking bottlenecks in web agents},
    author    = {Shlomov, Segev and Sela, Aviad and Levy, Ido and Galanti, Liane and Abitbol, Roy and others},
    journal   = {arXiv preprint arXiv:2409.01927},
    year      = {2024},
    url       = {https://arxiv.org/abs/2409.01927}
}

@article{yehudai2025survey,
    title     = {Survey on evaluation of {LLM}-based agents},
    author    = {Yehudai, Asaf and Eden, Lilach and Li, Alan and Uziel, Guy and Zhao, Yilun and Bar-Haim, Roy and Cohan, Arman and Shmueli-Scheuer, Michal},
    journal   = {arXiv preprint arXiv:2503.16416},
    year      = {2025},
    url       = {https://arxiv.org/abs/2503.16416}
}

@article{huang2024understanding,
    title     = {Understanding the planning of {LLM} agents: {A} survey},
    author    = {Huang, Xu and Liu, Weiwen and Chen, Xiaolong and Wang, Xingmei and Wang, Hao and Lian, Defu and Wang, Yasheng and Tang, Ruiming and Chen, Enhong},
    journal   = {arXiv preprint arXiv:2402.02716},
    year      = {2024},
    url       = {https://arxiv.org/abs/2402.02716}
}

@inproceedings{deng2023mind2web,
    title     = {{Mind2Web}: Towards a generalist agent for the web},
    author    = {Deng, Xiang and Gu, Yu and Zheng, Boyuan and Chen, Shijie and Stevens, Sam and Wang, Boshi and Sun, Huan and Su, Yu},
    booktitle = {Advances in Neural Information Processing Systems (NeurIPS)},
    volume    = {36},
    pages     = {28091--28114},
    year      = {2023},
    url       = {https://arxiv.org/abs/2306.06070}
}

@article{agashe2025agent,
    title     = {Agent {S2}: {A} Compositional Generalist-Specialist Framework for Computer Use Agents},
    author    = {Agashe, Saaket and Wong, Kyle and Tu, Vincent and Yang, Jiachen and Li, Ang and Wang, Xin Eric},
    journal   = {arXiv preprint arXiv:2504.00906},
    year      = {2025},
    url       = {https://arxiv.org/abs/2504.00906}
}

@article{marreed2025towards,
    title     = {Towards Enterprise-Ready Computer Using Generalist Agent},
    author    = {Marreed, Sami and Oved, Alon and Yaeli, Avi and Shlomov, Segev and Levy, Ido and Sela, Aviad and Adi, Asaf and Mashkif, Nir},
    journal   = {arXiv preprint arXiv:2503.01861},
    year      = {2025},
    url       = {https://arxiv.org/abs/2503.01861}
}

@inproceedings{zhang2025planning,
    title     = {Planning with Multi-Constraints via Collaborative Language Agents},
    author    = {Zhang, Cong and Goh, Xin Deik and Li, Dexun and Zhang, Hao and Liu, Yong},
    booktitle = "Proceedings of the 31st International Conference on Computational Linguistics",
    pages     = {10054--10082},
    year      = {2025},
    url       = {https://arxiv.org/abs/2405.16510}
}

@inproceedings{thakkarwebarena,
    title     = {{WebArena} Verified: Reliable Evaluation for Web Agents},
    author    = {Thakkar, Megh and Chapados, Nicolas and Pal, Christopher and others},
    booktitle = {NeurIPS 2025 Posters},
    year      = {2025},
    url       = {https://openreview.net/pdf?id=94tlGxmqkN}
}

@article{erdogan2025plan,
    title     = {Plan-and-act: Improving planning of agents for long-horizon tasks},
    author    = {Erdogan, Lutfi Eren and Lee, Nicholas and Kim, Sehoon and Moon, Suhong and Furuta, Hiroki and Anumanchipalli, Gopala and Keutzer, Kurt and Gholami, Amir},
    journal   = {arXiv preprint arXiv:2503.09572},
    year      = {2025},
    url       = {https://arxiv.org/abs/2503.09572}
}

@article{gur2023real,
    title     = {A real-world webagent with planning, long context understanding, and program synthesis},
    author    = {Gur, Izzeddin and Furuta, Hiroki and Huang, Austin and Safdari, Mustafa and Matsuo, Yutaka and Eck, Douglas and Faust, Aleksandra},
    journal   = {arXiv preprint arXiv:2307.12856},
    year      = {2023},
    url       = {https://arxiv.org/abs/2307.12856}
}

@misc{dechezelles2025browsergymecosystem,
    title     = {The {BrowserGym} Ecosystem for Web Agent Research}, 
    author    = {Thibault Le Sellier De Chezelles and Maxime Gasse and Alexandre Drouin and Massimo Caccia and Léo Boisvert and Megh Thakkar and Tom Marty and Rim Assouel and Sahar Omidi Shayegan and Lawrence Keunho Jang and Xing Han Lù and Ori Yoran and Dehan Kong and Frank F. Xu and Siva Reddy and Quentin Cappart and Graham Neubig and Ruslan Salakhutdinov and Nicolas Chapados and Alexandre Lacoste},
    year      = {2025},
    journal   = {arXiv preprint arXiv:2412.05467},
    url       = {https://arxiv.org/abs/2412.05467}
}

@article{chae2024web,
    title     = {Web Agents with World Models: Learning and Leveraging Environment Dynamics in Web Navigation},
    author    = {Chae, Hyungjoo and Kim, Namyoung and Ong, Kai Tzu-iunn and Gwak, Minju and Song, Gwanwoo and Kim, Jihoon and Kim, Sunghwan and Lee, Dongha and Yeo, Jinyoung},
    journal   = {arXiv preprint arXiv:2410.13232},
    year      = {2024},
    url       = {https://arxiv.org/abs/2410.13232}
}

@inproceedings{valmeekam2023planning,
    title     = {On the planning abilities of large language models-a critical investigation},
    author    = {Valmeekam, Karthik and Marquez, Matthew and Sreedharan, Sarath and Kambhampati, Subbarao},
    booktitle = {Advances in Neural Information Processing Systems (NeurIPS)},
    volume    = {36},
    pages     = {75993--76005},
    year      = {2023},
    url       = {https://arxiv.org/abs/2302.06706}
}

@inproceedings{xue2025illusion,
    title     = {An Illusion of Progress? Assessing the Current State of Web Agents},
    author    = {Tianci Xue and Weijian Qi and Tianneng Shi and Chan Hee Song and Boyu Gou and Dawn Song and Huan Sun and Yu Su},
    booktitle = {Proceedings of the Second Conference on Language Modeling (COLM)},
    year      = {2025},
    url       = {https://openreview.net/forum?id=6jZi4HSs6o}
}

@article{wei2025plangenllms,
    title     = {{PlanGenLLMs}: A modern survey of {LLM} planning capabilities},
    author    = {Wei, Hui and Zhang, Zihao and He, Shenghua and Xia, Tian and Pan, Shijia and Liu, Fei},
    journal   = {arXiv preprint arXiv:2502.11221},
    year      = {2025},
    url       = {https://arxiv.org/abs/2502.11221}
}

@inproceedings{shi2017world,
    title     = {World of bits: An open-domain platform for web-based agents},
    author    = {Shi, Tianlin and Karpathy, Andrej and Fan, Linxi and Hernandez, Jonathan and Liang, Percy},
    booktitle = {Proceedings of the 34th International Conference on Machine Learning (ICML)},
    pages     = {3135--3144},
    year      = {2017},
    url       = {http://proceedings.mlr.press/v70/shi17a.html}
}

@article{chen2021evaluating,
  title={Evaluating large language models trained on code},
  author={Chen, Mark and Tworek, Jerry and Jun, Heewoo and Yuan, Qiming and Pinto, Henrique Ponde De Oliveira and Kaplan, Jared and Edwards, Harri and Burda, Yuri and Joseph, Nicholas and Brockman, Greg and others},
  journal={arXiv preprint arXiv:2107.03374},
  year={2021},
  url={https://arxiv.org/abs/2107.03374}
}

@inproceedings{lu2024weblinx,
  title     = {{{WebLINX}: Real-World Website Navigation with Multi-Turn Dialogue}},
  author    = {Lu, Xing Han and Kasner, Zden{\v{e}}k and Reddy, Siva},
  booktitle = {{Proceedings of the 41st International Conference on Machine Learning (ICML 2024)}},
  pages     = {33007--33056},
  volume    = {235},
  series    = {Proceedings of Machine Learning Research},
  address   = {Vienna, Austria},
  month     = jul,
  year      = {2024},
  publisher = {PMLR},
  url       = {https://proceedings.mlr.press/v235/lu24e.html}
}

\appendix

\section{Results for Task Categorization}
\label{app:hard_set}
To construct the \textit{Hard} subset (Section~\ref{sec:methods}), we evaluated 5 backbone models over $R=5$ independent runs on all 381 WebArena test tasks. 
\begin{table}[h]
\centering
  \begin{tabular}{llll}
    \hline
    \textbf{Model}           & \textbf{Min} & \textbf{Max} & \textbf{Average} \\
    \hline
    \textbf{GPT 4o-mini}       &  16.3          &    20.7  & 17.9                       \\
    \textbf{GPT 4.1 mini}     & 24.1        &   33.9 & 29.0                        \\
    \textbf{GPT 4.1}       & 31.5           &   34.1 & 32.9                         \\
    \textbf{Qwen 2.5-VL 72B} &  22.8     &  23.9 & 23.5                         \\
    \textbf{Gemini 2.5 Flash} &  29.4     &  32.0 & 30.7                         \\
    \hline
  \end{tabular}
  \caption{
    Success Rate (SR) on test set of \textit{BrowserGym} on 5 baseline models.
  }
  \label{tab:hardset_generation}
\end{table}

Table~\ref{tab:hardset_generation} reports the resulting Success Rates (SR). The 3 strongest backbones: GPT-4.1, Qwen-2.5-VL and Gemini 2.5 achieve high success rate which motivated the selection done for the main experiments.

After running these experiments the division of task concluded in: 158 \textit{Hard} task, 14 \textit{Easy} tasks and 209 \textit{Medium} tasks. To better understand the distribution of the \textit{Hard}, \textit{Medium} and \textit{Easy} set we separated how the partition would be if only one model was considered.

\begin{table}[h]
\centering
 \resizebox{\linewidth}{!}{
  \begin{tabular}{llrrrr}
    \hline
    \textbf{Year of release} & \textbf{Model} & \textbf{Hard} & \textbf{Medium} &  \textbf{Easy} \\
    \hline
    July, 2024 & \textbf{GPT 4o-mini}   & 268 & 84 & 29  \\
    February, 2025 & \textbf{Qwen 2.5-VL 72B}      & 240 & 87 & 54  \\
    April, 2025 & \textbf{GPT 4.1-mini}  & 216 & 103 & 62  \\
    April, 2025 & \textbf{GPT 4.1}       & 213 & 81 & 87  \\   
    June, 2025 & \textbf{Gemini 2.5 Flash}    & 219 & 88 & 74 \\
    \hline
  \end{tabular}}
  \caption{
    Number of \textit{Hard}, \textit{Medium} and \textit{Easy} tasks per model.
  }
  \label{tab:subset_model}
\end{table}

In Table~\ref{tab:subset_model}, we can further appreciate the reason for the small number of \textit{Easy} tasks. Recall from Section~\ref{sec:methods}, that our definition for \textit{Easy} tasks require that all models solve the task in all runs \(R\); this strict definition results in only some of the tasks being in this category, meaning a task that is easy for one model is rarely also easy for another one. Additionally, this table shows that newer models (even with just a few months of difference) tend to perform better. This highlights that using a model as old as 4o-mini (released in July 2024) significantly lowers the number of easy tasks. For the interest of this project, studying the positive effects of planning, even an older model, is still interesting.

\subsection{Hard Task Distribution}
\begin{figure}[htp]
    \centering
    \includegraphics[width=\linewidth]{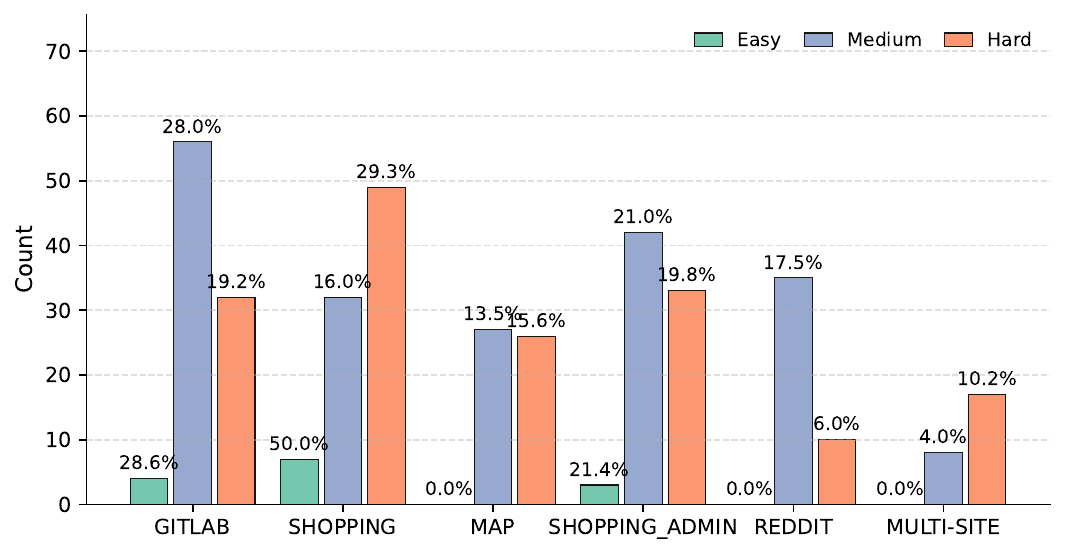}
    \caption{Distribution of \textit{Easy}, \textit{Medium}, and \textit{Hard} tasks across WebArena domains. Percentages indicate each domain’s share within a difficulty tier.}
    \label{fig:task_distribution}
\end{figure}

To better understand the nature of difficult tasks, Figure~\ref{fig:task_distribution} presents the distribution of task difficulty across domains. Intuitively, the \textit{Hard} subset contains most multi-site tasks, tasks that require cross-domain coordination and persistent memory across navigation steps. A substantial portion also comes from the Shopping and GitLab environments, this suggests that difficulty is closely tied to long-horizon reasoning as well as internal issues in these environments aligned to what is noted in~\cite{thakkarwebarena}.

\subsection{Task Categorization Stability Analysis}
To ensure that the \textit{Hard} subset was not merely an artifact of ``unlucky” runs, we performed a sensitivity analysis by recomputing the task partitions with different numbers of runs, from $R=1$ to $R=5$.

\begin{figure}[htp]
    \centering
    \includegraphics[width=\linewidth]{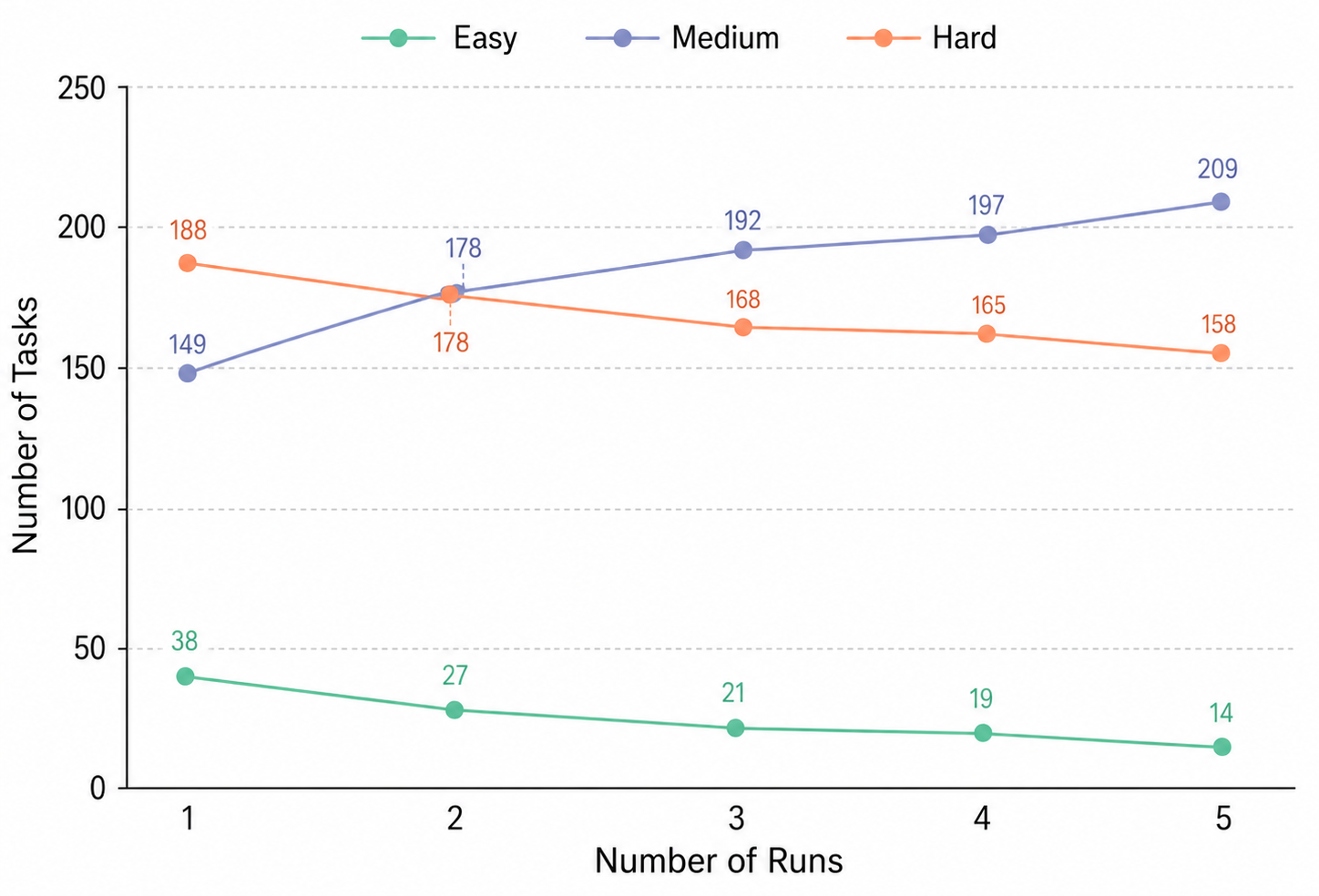}
    \caption{Task distribution across difficulty categories as the number of runs (R) increases.}
    \label{fig:task_dist}
\end{figure}

Figure~\ref{fig:task_dist} shows how the number of tasks in each difficulty category changes as $R$ increases. With a single run $(R=1)$ 188 tasks are classified as \textit{Hard}, however, as the number of runs increases, the size of the \textit{Hard} subset decreases to 158 tasks at $R=5$, while the \textit{Medium} subset increases from 149 to 209 tasks. This indicates that several tasks that appear unsolved under a single run become solvable at least once under repeated stochastic execution. This analysis leads to two main conclusions:

\begin{enumerate}
    \item Single-run evaluation can underestimate both task achievability and variability. Approximately 16\% of the tasks labeled \textit{Hard} under a single run are no longer classified as \textit{Hard} when evaluated over five runs. This suggests that some tasks are not consistently unsolvable (i.e. \textit{Hard}), but instead require repeated attempts or favorable stochastic trajectories to be solved at least once (i.e. \textit{Medium}). The opposite effect can also be observed in the \textit{Easy} set when a task solved in a single run is not necessarily solved reliably across repeated runs. This is reflected in the decreasing number of \textit{Easy} tasks as (R) increases, from 38 tasks at (R=1) to 14 tasks at (R=5). Together, these trends show that single-run evaluation can overestimate consistency while underestimating achievability.
    \item Multi-run grading improves reliability but does not guarantee convergence. Although overlap with the final \textit{Hard} set increases monotonically as the number of runs grows, Figure~\ref{fig:task_dist} does not demonstrate full convergence. The size of the \textit{Hard} subset continues to decrease as \(R\) increases (from 188 at \(R = 1\) to 158 at \(R = 5\)), suggesting that additional runs could further reduce the number of tasks classified as \textit{Hard}.
\end{enumerate}
In other words, the Hard subset should be interpreted as conditionally hard given $R = 5$ runs and the selected model family. This subset represents tasks that are consistently unsolved under these specific conditions, but not necessarily tasks that are fundamentally unsolvable. Despite this limitation, $R = 5$ already substantially reduces stochastic misclassifications compared to single-run grading. The resulting subset provides a sufficiently conservative and reproducible evaluation setting for studying whether planning can unlock tasks that non-planning agents consistently fail to solve.


\section{Strategy Sensitivity Results}
\label{app:strat_sensitivity}

Although absolute AR values on the Hard set seem small, bootstrap analysis reveals that these estimates can be highly stable (see Table \ref{tab:boot_AR} and \ref{tab:boot_STC}). It is important to note that the \textit{Hard} subset is defined as tasks for which all baseline agents fail in all runs, any \(AR>0\) already corresponds to solving tasks that a single agent (even a powerful model like GPT 4.1) could not solve. The bootstrap intervals show that, for the strategies with observed AR improvements, the entire confidence interval remains above zero, confirming that these gains are not explained by stochastic noise.

Importantly, AR and STC capture different aspects of variability: STC measures within-task reliability, while bootstrap CIs reflect uncertainty in the metric itself due to finite sampling. Using both, provides a clearer picture of agent robustness than SR alone, which treats all runs as independent and cannot distinguish “rarely solvable tasks” from “completely unsolved tasks.”

\begin{table}[h]
    \centering
    \resizebox{\linewidth}{!}{
    \begin{tabular}{c ll|llll}
    \hline
    \multicolumn{3}{c|}{\textbf{Configuration}} & \multicolumn{4}{c}{\textbf{}} \\
    \textbf{\#} & \textbf{Planner} & \textbf{Executor} & \textbf{Sequential} & \textbf{Checklist} & \textbf{Pseudocode} & \textbf{Narrative} \\
    \hline
        1 & GPT 4.1-mini & GPT 4.1-mini & 3.59 -- 6.59 & 1.20 -- 4.79 & 2.40 -- 4.79 & 5.39 -- 6.59 \\ 
        2 & Qwen 2.5-VL 72B & Qwen 2.5-VL 72B & 2.40 -- 2.99 & 2.40 -- 4.79 & 3.59 -- 3.59 & 2.99 -- 4.19 \\ 
        3 & Gemini 2.5 Flash & Gemini 2.5 Flash & 3.80 -- 5.70 & 5.06 -- 7.59 & 4.43 -- 8.23 & 6.33 -- 7.59 \\      
        4 & GPT 4.1-mini & Qwen 2.5-VL 72B & 2.99 -- 5.99 & 4.79 -- 8.38 & 2.40 -- 5.39 & 3.59 -- 5.39 \\ 
        5 & Qwen 2.5-VL 72B & GPT 4.1-mini & 2.99 -- 5.99 & 1.80 -- 4.19 & 1.80 -- 5.30 & 4.19 -- 6.59 \\
        6 & Qwen 2.5-VL 72B & Gemini 2.5 Flash & 5.70 -- 8.86 & 5.70 -- 6.96 & 3.80 -- 4.43 & 5.70 -- 8.86 \\
        7 & Gemini 2.5 Flash & Qwen 2.5-VL 72B & 1.27 -- 3.16 & 3.16 -- 5.70 & 1.90 -- 3.80 & 2.52 -- 4.43 \\
        8 & Gemini 2.5 Flash & GPT 4.1-mini & 4.43 -- 6.96 & 1.27 -- 3.16 & 2.53 -- 4.43 & 3.16 -- 5.70 \\
        9 & GPT 4.1-mini & Gemini 2.5 Flash & 6.96 -- 10.76 & 5.06 -- 8.86 & 4.43 -- 6.96 & 5.70 -- 9.49 \\
    \hline
    \end{tabular}
    }
    \caption{Bootstrap 95\% confidence intervals for AR on the \textit{Hard} subset with 1,000 resamples}
    \label{tab:boot_AR}
\end{table}

\begin{table}[h]
    \centering
    \small
    \resizebox{\linewidth}{!}{
    \begin{tabular}{c ll|llll}
    \hline
    \multicolumn{3}{c|}{\textbf{Configuration}} & \multicolumn{4}{c}{\textbf{}} \\
    \textbf{\#} & \textbf{Planner} & \textbf{Executor} & \textbf{Sequential} & \textbf{Checklist} & \textbf{Pseudocode} & \textbf{Narrative} \\
    \hline
        1 & GPT 4.1-mini & GPT 4.1-mini & 40.00 -- 72.00 & 32.50 -- 84.00 & 34.94 -- 70.00 & 70.91 -- 88.89 \\ 
        2 & Qwen 2.5-VL 72B & Qwen 2.5-VL 72B & 68.00 -- 96.00 & 30.00 -- 75.00 & 70.00 -- 93.33 & 57.14 -- 88.00 \\ 
        3 & Gemini 2.5 Flash & Gemini 2.5 Flash & 53.33 -- 80.00 & 58.33 -- 84.44 & 43.08 -- 77.14 & 68.33 -- 92.73 \\       
        4 & GPT 4.1-mini & Qwen 2.5-VL 72B & 34.29 -- 66.67 & 32.31 -- 67.50 & 35.56 -- 80.00 & 40.00 -- 70.00 \\ 
        5 & Qwen 2.5-VL 72B & GPT 4.1-mini & 28.00 -- 62.86 & 40.00 -- 76.00 & 30.00 -- 69.00 & 54.55 -- 87.50 \\
        6 & Qwen 2.5-VL 72B & Gemini 2.5 Flash & 49.18 -- 74.00 & 49.18 -- 74.00 & 68.57 -- 94.29 & 50.00 -- 76.00 \\
        7 & Gemini 2.5 Flash & Qwen 2.5-VL 72B & 26.67 -- 70.00 & 37.78 -- 75.00 & 36.67 -- 76.00 & 40.00 -- 73.33 \\
        8 & Gemini 2.5 Flash & GPT 4.1-mini & 45.45 -- 74.29 & 36.00 -- 70.00 & 45.71 -- 80.00 & 47.50 -- 76.03 \\
        9 & GPT 4.1-mini & Gemini 2.5 Flash & 44.29 -- 71.67 & 40.00 -- 68.57 & 53.55 -- 80.00 & 36.00 -- 69.09 \\
    \hline
    \end{tabular}
    }
    \caption{Bootstrap 95\% confidence intervals for STC on the \textit{Hard} subset with 1,000 resamples}
    \label{tab:boot_STC}
\end{table}

\section{Plan Representation Prompts}
\label{app:prompts}

The input of \method's planner consists of only a screenshot and the goal. Based on these two, the planner has to create a plan to complete the task. Each of the strategies proposed has their individual prompt but they all follow a similar format.

The original prompt was inspired from~\cite{agashe2025agent} and adapted to each planning strategy, having these prompts as results.

Strategy 1, sequential plan shown in Figure \ref{fig:sequential_prompt}.

Strategy 2, checklist shown in Figure \ref{fig:checklist_prompt}.

Strategy 3, pseudocode shown in Figure \ref{fig:pseudo_code}.

Strategy 4, narrative shown in Figure \ref{fig:narrative_prompt}.

\section{Single-Agent Plan Creation Failure Cases}
\label{app:dynamic_plans}
When a single agent both produces the plan and predicts the next action it can sometimes get confused at the step it is currently at and start repeating the same action entering what we call an \textit{action loop} or get over confident thinking a step was completed when it was not. Some examples are shown below.

\subsection{Shopping Admin Case}
\begin{figure*}[t]
    \centering
    \begin{subfigure}[b]{0.45\textwidth}
        \centering
        \includegraphics[width=\textwidth]{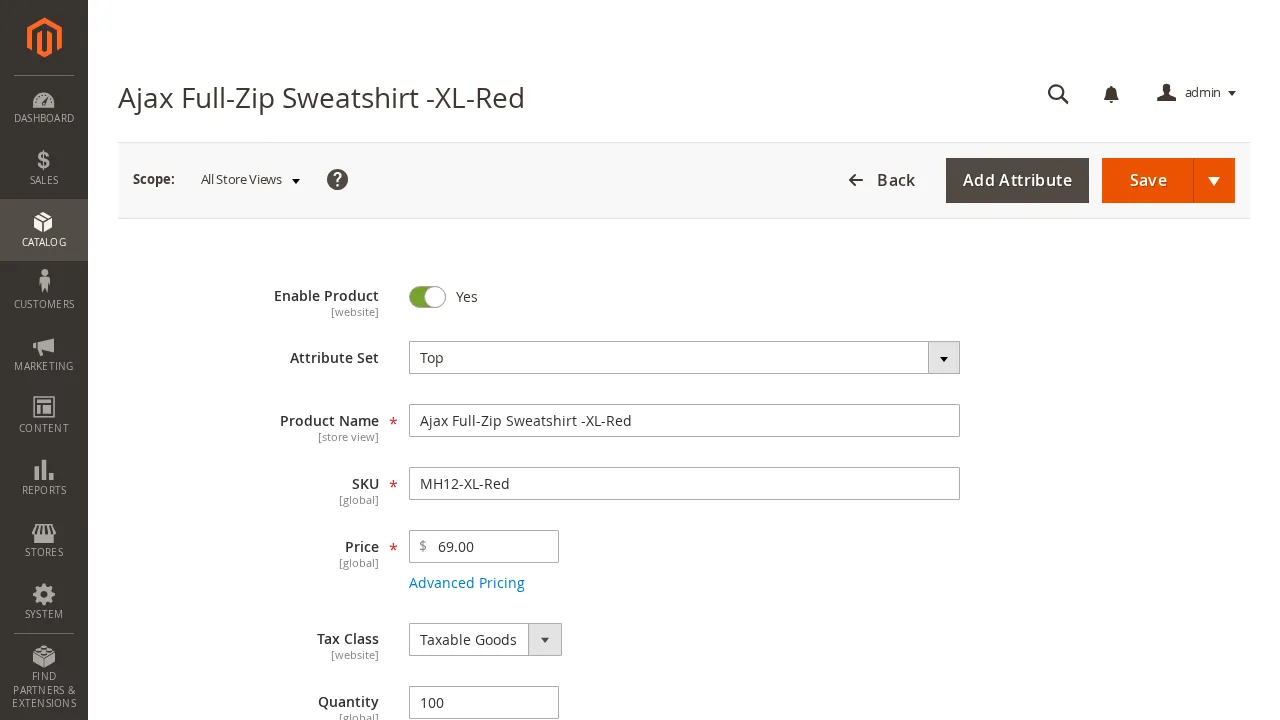}
        \caption{Step 0}
        \label{img:step0_0}
    \end{subfigure}
    \hfill
    \begin{subfigure}[b]{0.45\textwidth}
        \centering
        \includegraphics[width=\textwidth]{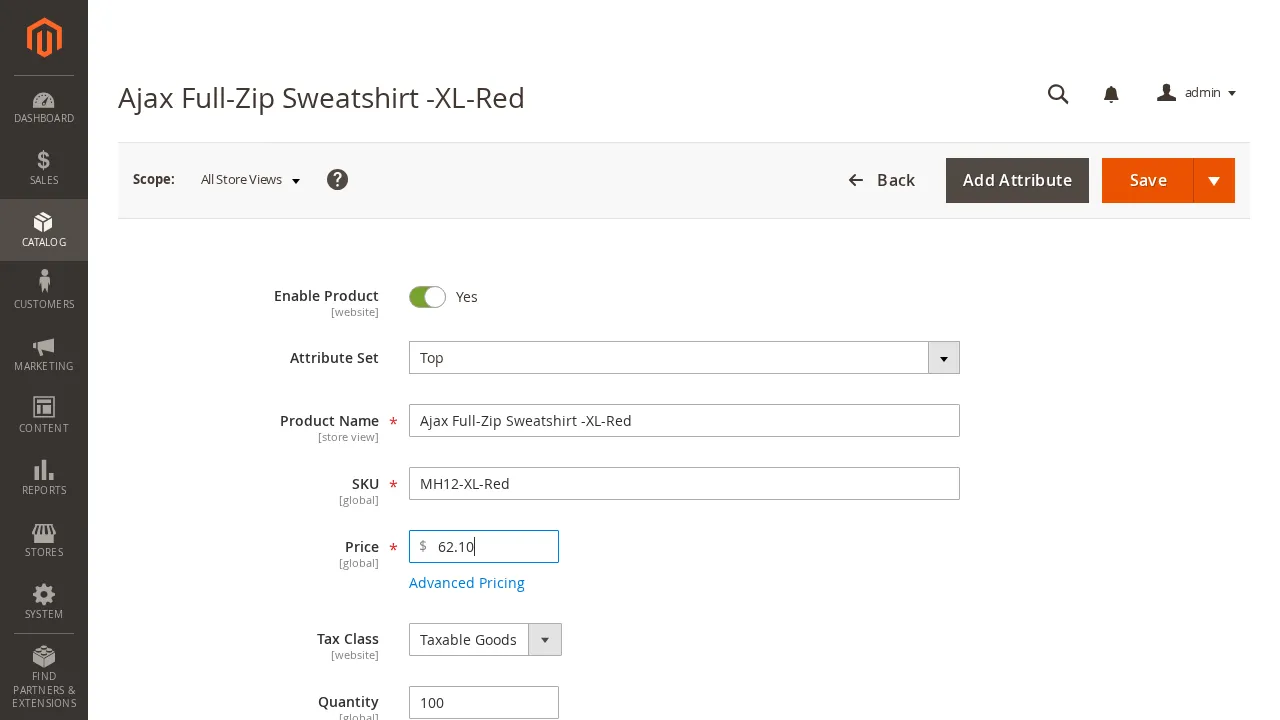}
        \caption{Step 1}
        \label{img:step1_0}
    \end{subfigure}

    \vspace{0.5cm}

    \begin{subfigure}[b]{0.45\textwidth}
        \centering
        \includegraphics[width=\textwidth]{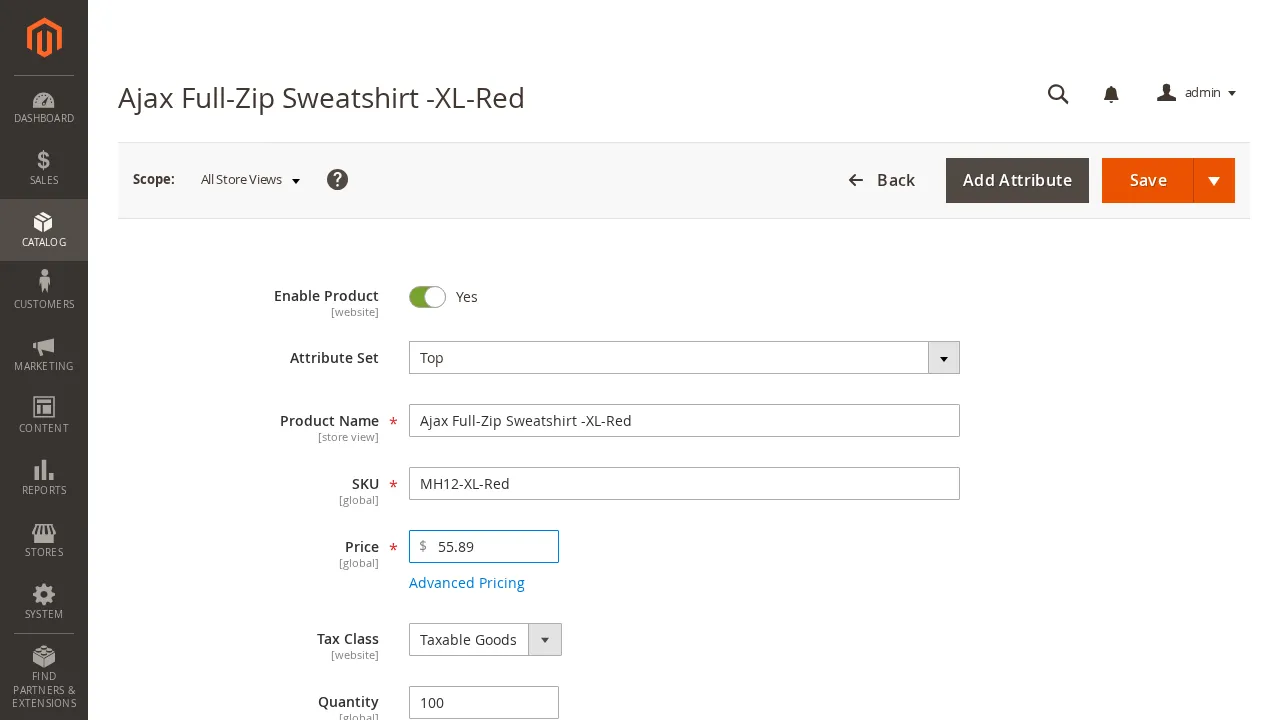}
        \caption{Step 2}
        \label{img:step2_0}
    \end{subfigure}
    \hfill
    \begin{subfigure}[b]{0.45\textwidth}
        \centering
        \includegraphics[width=\textwidth]{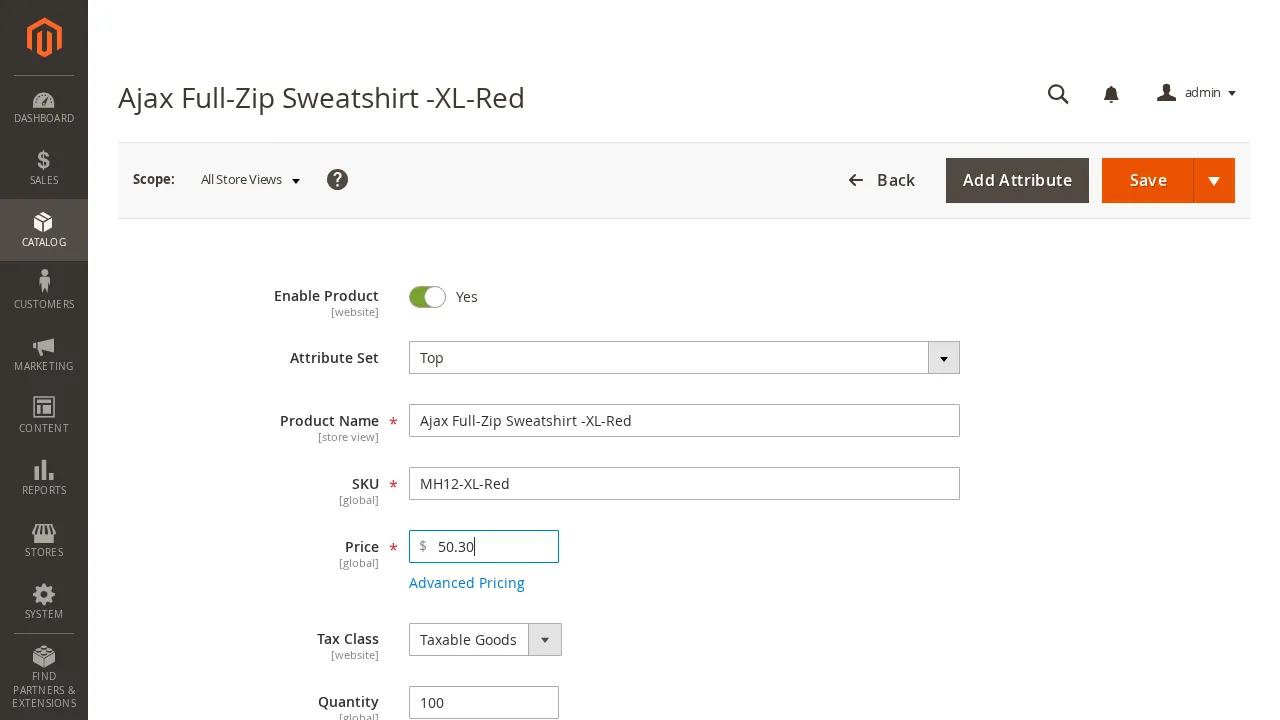}
        \caption{Step 3}
        \label{img:step3_0}
    \end{subfigure}

    \caption{Example of single-agent planner-executor on task webarena.459. The agent incorrectly keeps reducing the price.}
    \label{fig:combined_webarena}
\end{figure*}

The task webarena.459: "Reduce the price of this product by 10\%". The task is initiated at the product page in the shopping admin website as shown in Figure \ref{img:step0_0}.

At the next step the agent correctly calculates the 10\% and reduces the price (see Figure \ref{img:step1_0}).

However, instead of saving the reduced priced, it proceeds to reduce the price again, see Figure \ref{img:step2_0} and again, see Figure \ref{img:step3_0} and keeps reducing the price until step 30. A similar behaviour happens with task webarena.462 which is to increase the price.

\subsection{Map Case}
Task webarena.54 states "How long does it take to walk from Carnegie Mellon University to Univ of Pittsburgh?"


In this task the agent starts in the correct manner picking the right origin, destination and form of travel. However, at the 4th step when the agent is supposed to click go. The plan in the prompt is as following:
\begin{tcolorbox}
\begin{verbatim}
Plan:

You just executed step 3 of the 
previously proposed plan:
1. Change the mode of transportation to 
"Foot (OSRM)".
2. Click the "Go" button to get walking 
directions.
3. Extract the walking time from the 
results.

After reviewing the effect of your 
previous actions, verify if your plan
is still relevant and update it 
if necessary.
\end{verbatim}
\end{tcolorbox}
As shown  in the plan above, the step of `Clicking Go' is ignored and is assumed completed. The agent proceeds to just "wait" for the results to appear instead of trying to click go again it proceeds to just wait (noop) for the remaining 30 turns.

\subsection{Multiple Runs are Important}

 As shown in Tables~\ref{tab:main_table_AR} and~\ref{tab:main_table_SR}, relying solely on Success Rate (SR) can obscure meaningful differences. SR can fail to indicate how many tasks an agent can actually solve, whereas AR and STC together capture more directly both achievability and robustness. Given the stochastic nature of LLM-based agents, metrics that incorporate multiple runs are essential.
\begin{table*}[h]
\centering
\resizebox{\linewidth}{!}{
\begin{tabular}{c ll|cc|cc|cc|cc|cc}
\hline
\multicolumn{3}{c|}{\textbf{Configuration}} 
& \multicolumn{2}{c|}{\textbf{Dynamic}} 
& \multicolumn{8}{c}{\textbf{Static}} \\

\textbf{\#} & \textbf{Planner} & \textbf{Executor}
& \multicolumn{2}{c|}{\textbf{Sequential}}
& \multicolumn{2}{c|}{\textbf{Sequential}}
& \multicolumn{2}{c|}{\textbf{Checklist}}
& \multicolumn{2}{c|}{\textbf{Pseudocode}}
& \multicolumn{2}{c}{\textbf{Narrative}} \\
\hline
&&& \textbf{SR} & \textbf{SE}
& \textbf{SR} & \textbf{SE}
& \textbf{SR} & \textbf{SE}
& \textbf{SR} & \textbf{SE}
& \textbf{SR} & \textbf{SE} \\
\hline

1 & GPT 4.1-mini & GPT 4.1-mini
& 2.1 & 0.5
& 2.1 & 0.6
& 1.5 & 0.4
& 1.5 & 0.5
& \textbf{3.8} & \textbf{0.7} \\

2 &  Qwen 2.5-VL 72B & GPT 4.1-mini
& - & -
& 1.7 & 0.5
& 1.7 & 0.4
& 1.7 & 0.4
& \textbf{4.0} & \textbf{0.7} \\

3 & Gemini & GPT 4.1-mini
& - & -
& \textbf{3.1} & \textbf{0.6}
& 1.1 & 0.4
& 2.2 & 0.5
& 3.0 & 0.6 \\

4 &  Qwen 2.5-VL 72B &  Qwen 2.5-VL 72B
& 2.8 & 0.6
& 2.1 & 0.5
& \textbf{3.8} & \textbf{0.4}
& 2.6 & 0.6
& 2.4 & 0.6 \\

5 & GPT 4.1-mini &  Qwen 2.5-VL 72B
& - & -
& 1.8 & 0.5
& 2.2 & 0.6
& 2.2 & 0.5
& \textbf{2.3} & \textbf{0.5} \\

6 & Gemini &  Qwen 2.5-VL 72B
& - & -
& 1.0 & 0.4
& \textbf{2.3} & \textbf{0.5}
& 1.5 & 0.4
& 1.8 & 0.5 \\

7 & Gemini 2.5 Flash & Gemini 2.5 Flash
& 3.7 & 0.8
& 3.0 & 0.6
& 4.4 & 0.7
& 3.5 & 0.7
& \textbf{5.6} & \textbf{0.8} \\

8 & GPT 4.1-mini & Gemini 2.5 Flash
& - & -
& \textbf{5.1} & \textbf{0.8}
& 3.9 & 0.7
& 3.7 & 0.7
& 4.1 & 0.7 \\

9 &  Qwen 2.5-VL 72B & Gemini 2.5 Flash
& - & -
& \textbf{4.5} & \textbf{0.7}
& 4.1 & 0.7
& 3.4 & 0.6
& \textbf{4.5} & \textbf{0.7} \\

\hline
\end{tabular}
}
\caption[Comparison of Success Rate (SR) and Standard Error (SE)]{
   Comparison of Success Rate (SR) and Standard Error (SE) across planning strategies on the \textit{Hard} subset}
\label{tab:main_table_SR}
\end{table*}

\begin{table*}[ht!]
\centering
\resizebox{\linewidth}{!}{
\begin{tabular}{c ll|rr|rr|rr|rr}
\hline
\multicolumn{3}{c|}{\textbf{Configuration}}
& \multicolumn{2}{c|}{\textbf{Sequential}}
& \multicolumn{2}{c|}{\textbf{Checklist}}
& \multicolumn{2}{c|}{\textbf{Pseudocode}}
& \multicolumn{2}{c}{\textbf{Narrative}} \\
\textbf{\#} & \textbf{Planner} & \textbf{Executor}
& \textbf{AR} & \textbf{STC}
& \textbf{AR} & \textbf{STC}
& \textbf{AR} & \textbf{STC}
& \textbf{AR} & \textbf{STC} \\
\hline
1 & GPT 4.1-mini & GPT 4.1-mini
& 100 & 100
& 100 & 100
& 100 & 100
& 100 & 100 \\
2 & Qwen 2.5-VL 72B & Qwen 2.5-VL 72B
& 100 & 100
& 100 & 100
& 100 & 96.0
& 85.7 & 98.0 \\
3 & Gemini 2.5 Flash & Gemini 2.5 Flash
& 100 & 100
& 100 & 100
& 100 & 100
& 100 & 100 \\
\bottomrule
\end{tabular}
}
\caption{Achievement Rate (AR) and Solved-Task Consistency (STC) of different strategies on the \textit{Easy} subset by model}
\label{tab:gen_agents_easy}
\end{table*}
\section{Additional Experiments and Results}


%


\subsection{Easy Set Runs}
\label{app:easy_set}

As a sanity check, we evaluate the \textit{Easy} subset to confirm that planning does not degrade performance on tasks that are already solvable. Experiments on the \textit{Easy} subset (see Table~\ref{tab:gen_agents_easy}) revealed that low-level plans can limit exploration and generalization. Executors sometimes failed when constrained to search for exact words or sections, even though more flexible reasoning would have sufficed. High-level guidance appears crucial for balancing structure with adaptability.

\begin{figure*}[t]
\centering  
\begin{tcolorbox}
You are an expert planning agent for solving GUI navigation tasks. Using the plan that you output, another language model will execute the task. 

You are provided with: 

1. The state of the computer screen through a desktop screenshot Your responsibilities: 

1. Generate a new plan to complete the task by decomposing the task into meaningful subtasks of logically ordered chunks. 

2. Ensure the plan is concise and contains only necessary steps. 

3. Carefully observe and understand the current state of the computer before generating your plan. 

4. Avoid including steps in your plan that the task does not ask for. 

5. Generate a thought process or explanation about the plan generated. 

Below are important considerations when generating your plan: 

1. Provide the plan in a step-by-step numbered format with sequential subtasks. 

2. Provide high-level subtasks and avoid over-specifying actions—this isn't a low-level how-to (e.g. avoid specifying mouse clicks). This is a logical task breakdown that captures essential steps and decisions. The executor will identify the fine-grained actions to be taken whilst interacting with the GUI using your subgoals. 

3. Do not include verification steps in your planning. Steps that confirm or validate other subtasks should not be included. 

4. Do not include optional steps in your planning. Your plan must be as concise as possible. 

5. Do not include unnecessary steps in your planning. If you are unsure if a step is necessary, do not include it in your plan. 

6. Your plan **must** be end-to-end meaning it should entail instructions from the initial state till the goal state. 

Below is an example of a plan created using subtask decomposition, given the request "Book a round-trip flight from New York to San Francisco for next weekend, departing Friday and returning Sunday, on the cheapest available airline.": 

1. Open a flight booking website (e.g., Google Flights, Expedia, etc.) 

2. Input departure city as New York and destination as San Francisco 

3. Set departure date to next Friday and return date to next Sunday 

4. Search for round-trip flights 

5. Sort results by price 

6. Select the cheapest available round-trip option that fits the criteria 

7. Proceed to booking and complete the purchase Important formatting guidelines: 

1. Make sure the formatting of your answer is correct. It should follow the following HTML tags format <observation> Description and understanding of current state </observation> <plan> Generated Plan </plan> <thought> Thought process </thought> 

2. Make sure to provide the correct start and end tags
\end{tcolorbox}
\caption{Prompt for sequential plans.}
\label{fig:sequential_prompt}
\end{figure*}
\begin{figure*}[t]
\begin{tcolorbox}
You are an expert planning agent for solving GUI navigation tasks. Using the plan that you output, another language model will execute the task. You are provided with: 1. The state of the computer screen through a desktop screenshot 

Your responsibilities: 

1. Generate a checklist of all necessary requirements that need to completed in order to achieve the goal. 

2. Ensure this checklist is concise and contains only necessary requirements 

3. Carefully observe and understand the current state of the computer before generating your checklist

4. Generate a thought process or explanation about the checklist generated 

Below are important considerations when generating your checklist: 

1. Provide the plan in a unordered checklist format. This is not necessarily sequential, but outlines requirements that must be met. The executor will identify the order of fine-grained actions to be taken whilst interacting with the GUI based on the provided checklist. 

2. Do not include verification steps. Steps that confirm or validate other subtasks should not be included. 

3. Do not include optional requirements. Your checklist must be as concise as possible. 

4. Do not include unnecessary requirements. If you are unsure if a requirement is necessary, do not include it in your checklist. 

Below is a example of a checklist given the request "Sign up for the weekly newsletter on the website.":

[ ] The email input field is filled with a valid email address. 

[ ] The "Subscribe" or "Sign Up" button has been clicked. 

[ ] A confirmation message or success banner is visible after submission. 

[ ] The user is not required to solve a CAPTCHA. 

[ ] There are no visible error messages on the page. 

[ ] The checkbox for accepting terms (if present) is checked. 

Important formatting guidelines: 

1. Make sure the formatting of your answer is correct. It should follow the following HTML tags format <observation> Description and understanding of current state </observation> <plan> Generated Checklist </plan> <thought> Thought process </thought> 

2. Make sure to provide the correct start and end tags
\end{tcolorbox}
\caption{Prompt for checklist plans.}
\label{fig:checklist_prompt}
\end{figure*}

\begin{figure*}[t]
\begin{tcolorbox}
You are an expert agent in web navigation and you are able to foresee multiple possible trajectories to achieve a navigation task. Using the plan that you output, another language model will execute the task. 

You are provided with: 

1. The state of the computer screen through a desktop screenshot 

Your responsibilities: 

1. Generate an algorithm style plan in order to achieve the goal in a pseudocode style with abstract functions depicting high-level actions. 

2. Ensure this pseudocode plan is as concise as possible avoiding complex nested if statements. 

3. Carefully observe and understand the current state of the computer before generating your pseudocode 

4. Generate a thought process or explanation about the pseudocode plan generated Below are important considerations when generating your pseudocode: 

1. Provide the pseudocode plan in a readable format. Make function names intuitive and descriptive. The executor will then determine the fine-grained actions to take using these function names. 

2. Consider potential obstacles or layout variations the executor may encounter whilst completing the task. 

3. Do not include verification sections. Steps that confirm or validate other subtasks should not be included. 

4. Do not include unnecessary parts in your pseudocode. If you are unsure if a requirement is necessary, do not include it. 

Below is an example of the correct formatting and level of abstraction for the request "Add red coat product to cart and wish list":
\begin{verbatim}
while(red_coat_not_in_cart and red_coat_not_in_wish_list) 
    search_for_red_coat 
    if item found: 
        click add to cart 
        if item in cart: 
            find_and_click_add_product_to_wishlist 
            if item_added_to_wishlist: 
                break 
            else: 
                find_add_to_wishlist_button 
        else: 
            find_add_to_cart_button 
    else: 
        update search query 
\end{verbatim}

Important formatting guidelines: 

1. Make sure the formatting of your answer is correct. It should follow the following HTML tags format <observation> Description and understanding of current state </observation> <plan> Generated Pseudocode Plan </plan> <thought> Thought process </thought> 

2. Make sure to provide the correct start and end tags
\end{tcolorbox}
\caption{Prompt for algorithmic pseudocode plans.}
\label{fig:pseudo_code}
\end{figure*}

\begin{figure*}[t]
\begin{tcolorbox}
You are an expert planning agent for solving GUI navigation tasks. Using the plan that you output, another language model will execute the task. You are provided with: 1. The state of the computer screen through a desktop screenshot Your responsibilities: 

1. Generate a paragraph describing what the agent should do in order to achieve the given goal. 

2. Ensure this paragraph is concise and contains only necessary information and suggestions to help the agent achieve the goal. 

3. Carefully observe and understand the current state of the computer before generating your paragraph 

4. Generate a thought process or explanation about the paragraph generated 

Below are important considerations when generating your paragraph: 

1. Provide the paragraph in plain text. There is no need for stylistic formatting. 

2. Do not include verification steps. Steps that confirm or validate other subtasks should not be included. 

3. Do not include unnecessary requirements. If you are unsure if a requirement is necessary, do not include it in your checklist. 

Below is an example of a paragraph describing how an executor agent should fulfill the request 

"Book a round-trip flight from New York to San Francisco for next weekend, departing Friday and returning Sunday, on the cheapest available airline.": To book a round-trip flight from New York to San Francisco for next weekend, you would start by opening a flight booking site like Google Flights or Expedia. Input New York as the departure city and San Francisco as the destination, then set the departure for Friday and the return for Sunday. After initiating the search, sort the results by price to find the cheapest option. Once identified, select the flight and proceed to the booking page to enter passenger and payment details. Finally, complete the purchase, ensuring no obstacles arise during the process. 

Important formatting guidelines: 

1. Make sure the formatting of your answer is correct. It should follow the following HTML tags format <observation> Description and understanding of current state </observation> <plan> Generated Paragraph </plan> <thought> Thought process </thought> 

2. Make sure to provide the correct start and end tags
\end{tcolorbox}
\caption{Prompt for narrative plans.}
\label{fig:narrative_prompt}
\end{figure*}

\end{document}